\documentclass[11pt,a4paper]{article}
\usepackage[hyperref]{emnlp-ijcnlp-2019}
\usepackage{times}
\usepackage{latexsym}
\usepackage{multirow}
\usepackage[para]{threeparttable}
\usepackage{CJKutf8}
\usepackage{censor}
\usepackage{graphicx}
\usepackage{cprotect}
\usepackage{hyperref}

\usepackage{url}

\aclfinalcopy 

\newcommand{\astfootnote}[1]{
    \let\oldthefootnote=\thefootnote
    \setcounter{footnote}{1}
    \renewcommand{\thefootnote}{\fnsymbol{footnote}}
    \footnotetext{#1}
    \let\thefootnote=\oldthefootnote
}

\title{Don't Just Scratch the Surface: \\ Enhancing Word Representations for Korean with Hanja}

\author{Kang Min Yoo$^*$, Taeuk Kim$^*$ \and Sang-goo Lee \\
  Department of Computer Science and Engineering \\
  Seoul National University, Seoul, Korea \\
  {\tt \{kangminyoo,taeuk,sglee\}@europa.snu.ac.kr} \\}
\date{}

\begin{document}
\maketitle
\begin{abstract}
  We propose a simple yet effective approach for improving Korean word representations using additional linguistic annotation (i.e. Hanja).
  We employ cross-lingual transfer learning in training word representations by leveraging the fact that Hanja is closely related to Chinese.
  We evaluate the intrinsic quality of representations learned through our approach using the word analogy and similarity tests.
  In addition, we demonstrate their effectiveness on several downstream tasks, including a novel Korean news headline generation task.
\end{abstract}

\section{Introduction}

\astfootnote{Equal contribution.}
\setcounter{footnote}{0}

There is a strong connection between the Korean and Chinese languages due to cultural and historical reasons \cite{lee2011history}.
Specifically, a set of logograms with very similar forms to the Chinese characters, called \textbf{Hanja}\footnote{Hanja and traditional Chinese characters are very similar but not completely identical. Some differences in strokes and language-exclusive characters exist.}, served in the past as the only medium for written Korean until \textbf{Hangul}, the Korean alphabet, and \textbf{Jamo} came into existence in 1443 \cite{sohn2001korean}. 
Considering this etymological background, a substantial portion of Korean words are classified as Sino-Korean, a set of Korean words that had originated from Chinese and can be expressed in both Hanja and Hangul \cite{taylor1997psycholinguistic}, with the latter becoming commonplace in modern Korean.

\begin{figure}[t!]
    \centering
    \includegraphics[width=1.0\linewidth]{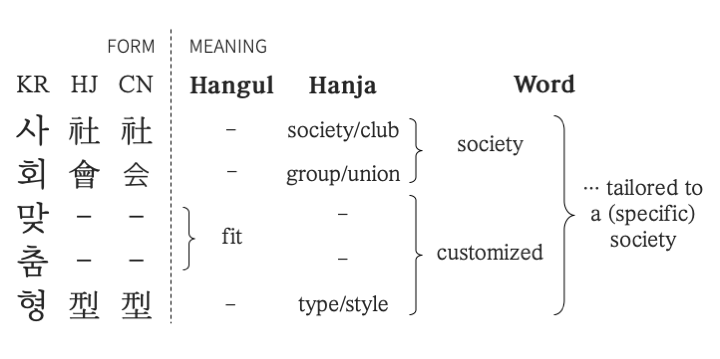}
    \cprotect\caption{An example of a Korean word showing its form and multi-level meanings. The Sino-Korean word consists of Hangul phonograms (\verb|KR|) and Hanja logograms (\verb|HJ|). Although annotation of Hanja is optional, it offers deeper insight into the word meaning due to its association with the Chinese characters (\verb|CN|).}
    \label{fig:example}
\end{figure}

Based on these facts, we assume the introduction of Hanja-level information will aid in grounding better representation for the meaning of Korean words (e.g. see Fig. \ref{fig:example}). 
To validate our hypothesis, we propose a simple yet effective approach to training Korean word representations, named \textbf{Hanja-level SISG} (Hanja-level Subword Information Skip-Gram), for capturing the semantics of Hanja and subword structures of Korean and introducing them into the vector space.
Note that it is also quite intuitive for native Koreans to resolve the ambiguity of (Sino-)Korean words with the aid of Hanja.
We conjecture this heuristic is attributed to the fact that Hanja characters are logograms, each of which contains more lexical meaning when compared against its counterpart, the Hangul character, which is a phonogram.
Accordingly, in this work, we focus on exploiting the rich extra information from Hanja as a means of constructing better Korean word embeddings. 
Furthermore, our approach enables us to empirically investigate the potential of character-level cross-lingual knowledge transfer with a case study of Chinese and Hanja.

To sum up, our contributions are threefold:

\begin{itemize}
    \item We introduce a novel way of training Korean word embeddings with an expanded Korean alphabet vocabulary using Hanja in addition to the existing Korean characters, Hangul. 
    \item We also explore the possibility of character-level knowledge transfer between two languages by initializing Hanja embeddings with Chinese character embeddings before training skip-gram.
    \item We prove the effectiveness of our method in two ways, one of which is intrinsic evaluation (the word analogy and similarity tests), and the other being two downstream tasks including a newly proposed Korean news headline generation task.
\end{itemize}

\section{Model Description}
\label{sec:model}

\subsection{Skip-Gram (SG)} 
Given a corpus as a sequence of words $\left( w_1, w_2, \ldots , w_T \right)$, the goal of a skip-gram model \cite{mikolov2013distributed} is to maximize the log-probabilities of context words given a target word $w_i$:

\begin{equation}
    \sum_{i \in \left\{ 1, \ldots, T \right\}} \sum_{c \in \mathcal{C}\left(i\right)} {\log{p \left( w_c \middle| w_i \right)}},
\end{equation}
where $\mathcal{C}$ returns a set of context word indices given a word index. The usual choice for training the parameterized probability function $p \left( w_c \middle| w_i \right)$ is to reframe the problem as a set of binary classification tasks of the target context word $c$ and other negative samples chosen from the entire dictionary (negative sampling). The objective is thus

\begin{equation}
 L \left( s \left( w_i, w_c \right) \right) + \sum_{j \in \mathcal{N} \left( i, c \right)} L \left( -s \left( w_i, w_j \right) \right),
\end{equation}
where $L \left( x \right) = \log { \left( 1 + e ^ {-x} \right) }$, the binary logistic loss. $\mathcal{N}$ returns a set of negative sample indices given target word and context indices. In the vanilla SG model, the scoring function $s$ is the dot product between the two word vectors: $\mathbf{v}_w$ and $\mathbf{v}_c$. 

\subsection{Subword Information SG (SISG) and Jamo-level SISG} 
In \citet{bojanowski2017enriching}, the scoring function incorporates subword information. Specifically, given all possible n-gram character sets $\mathbf{G} = \left\{g_1, \ldots, g_G\right\}$ in the corpus and that the n-gram set of a particular word $w$ is $\mathcal{G}_w \subset \mathbf{G}$, the scoring function is thus the sum of dot products between all n-gram vectors $\mathbf{z}$ and the context vector $\mathbf{v}_c$:

\begin{equation}
    s \left( w, c \right) = \sum_{g \in \mathcal{G}_w} {\mathbf{z}_g \mathbf{v}_c}.
    \label{eq:subword}
\end{equation}

However, due to the deconstructible structure of Korean characters and the agglutinative nature of the language, the model proposed by \citet{bojanowski2017enriching} comes short in capturing sub-character and inter-character information specific to Korean understanding. As a remedy, the model proposed by \citet{park2018subword} introduces jamo-level n-grams $g^{(j)}$, whose vectors can be associated with the target context as well. Given that a Korean word $w$ can be split into a set of jamo n-grams $\mathcal{G}_w^{(j)} \subset \mathbf{G}^{(j)} = \left\{ g^{(j)}_1, \ldots, g^{(j)}_J \right\}$, where $\mathbf{G}^{(j)}$ is the set of all possible jamo n-grams in the corpus, the updated scoring function for jamo-level skip-gram $s^{(j)}$ is thus

\begin{equation}
    s^{(j)} \left( w, c \right) = s \left( w, c \right) + \sum_{j \in \mathcal{G}_w^{(j)}} {\mathbf{z}_j \mathbf{v}_c}.
    \label{eq:jamo}
\end{equation}

\subsection{Hanja-level SISG\footnote{The code and models are publicly available online at \url{https://github.com/kaniblu/hanja-sisg}}}
Another semantic level in Korean lies in Hanja logograms. Hanja characters can be mapped to Hangul phonograms, hence Hanja annotations are sufficient but not necessary in conveying meanings (Fig. \ref{fig:example}). 
As each Hanja character is associated with semantics, their presence could provide meaningful aid in capturing word-level semantics in the vector space. 

We propose incorporating Hanja n-grams in the learning of word representations by allowing the scoring function to associate each Hanja n-gram with the target context word. Concretely, given that a Korean word $w$ contains a set of Hanja sequences $\mathcal{H}_w = \left( h_{w,1}, \ldots, h_{w, H_w} \right)$ (e.g. for the example in Fig. \ref{fig:example} $h_{w,1} = $ ``\begin{CJK}{UTF8}{bkai}\small社會\end{CJK}" and $h_{w,2} = $ ``\begin{CJK}{UTF8}{bkai}\small型\end{CJK}") and that $\mathcal{I}_h$ is the set of Hanja n-grams for a Hanja sequence $h$ (e.g. $\mathcal{I}_{h_{w,2}} = $ \{``\begin{CJK}{UTF8}{bkai}\small型\end{CJK}", ``\verb|<boh>|\begin{CJK}{UTF8}{bkai}\small型\end{CJK}", ``\begin{CJK}{UTF8}{bkai}\small型\end{CJK}\verb|<eoh>|"\}, when n is equal to 2), \textit{all} Hanja n-grams $\mathcal{G}^{(h)}_w$ for the word $w$ are the union of Hanja n-grams of Hanja sequences present in $w$: $\bigcup_{h \in \mathcal{H}_w} \mathcal{I}_h$. The length of Hanja n-grams in $\mathcal{I}_h$ is a hyperparameter. Then the new score function for Hanja-level SISG is

\begin{equation}
    s^{(h)} \left( w, c \right) = s^{(j)} \left( w, c \right) + \sum_{h \in \mathcal{G}_w^{(h)}} {\mathbf{z}_h \mathbf{v}_c},
    \label{eq:Hanja}
\end{equation}
where $\mathbf{z}_h$ is the n-gram vector for a Hanja sequence $h$.

\subsection{Character-level Knowledge Transfer} 
Since Hanja characters essentially share deep roots with Chinese characters, many of them can be mapped by one-to-one correspondence. By converting Korean characters into simplified Chinese characters or vice versa, it is possible to utilize pre-trained Chinese embeddings in the process of training Hanja-level SISG. In our case, we propose leveraging advances in Chinese word embeddings from the Chinese community by initializing Hanja n-grams vectors $\mathbf{z}_h$ with the state-of-the-art Chinese embeddings \cite{li2018analogical} and training with the score function in Equation \ref{eq:Hanja}.

\section{Experiments}

\subsection{Word Representation Learning}

\subsubsection{Korean Corpus} 
For learning word representations, we utilize the corpus prepared by \citet{park2018subword} with small modifications.
We perform additional data cleansing (e.g. removing non-Korean sentences in the corpus and unifying number tags) to obtain a cleaner version of the corpus. All of our comparative studies are based on this corpus. 
For training Hanja-level SISG, we use Hanjaro\footnote{\href{http://hanjaro.juntong.or.kr/}{http://hanjaro.juntong.or.kr/}} tagger to automatically annotate all of our datasets with Hanja. 
It is the state-of-the-art Hanja tagger available to the public to the best of our knowledge.

\subsubsection{Models}
We compare our approach with three other baselines. 
Skip-Gram (\verb|SG|) model \cite{mikolov2013distributed} does not incorporate any n-gram information in the training loss. 
Subword Information Skip-Gram (\verb|SISG(c)|) \cite{bojanowski2017enriching} uses character-level n-gram information to enrich word representations. 
For the Korean language, character-level n-grams correspond to syllable n-grams. 
Jamo-level SISG (\verb|SISG(cj)|) \cite{park2018subword} uses jamo-level n-grams in addition to character-level n-grams. 
Our ablation studies confirm that the hyperparameter settings (character n-grams ranging from 1 to 6 and jamo n-grams ranging from 3 to 5) proposed by the authors are indeed optimal, hence our subsequent studies all employ the same settings. 
Our approach is denoted by \verb|SISG(cjh)|). 
As for the specific hyperparameter settings of our approach, \verb|SISG(cjh3)| denotes Hanja n-grams ranging from 1 to 3 while \verb|SISG(cjh4)| denotes Hanja n-grams ranging from 1 to 4. 
\verb|SISG(cjhr)| is a special version of our approach, where Hanja n-grams are randomly initialized as opposed to being pre-trained (Section 2.4). 
The details of our ablation studies and the model implementation are included in the supplementary material.

\begin{table}[t!]
\small
\begin{center}
\begin{threeparttable}
    \begin{tabular}{|l|ccc|cc|}
    \hline \multirow{2}{*}{\textbf{Method}} & \multicolumn{3}{c|}{\textbf{Analogy}} & \multicolumn{2}{c|}{\textbf{Similarity}}\\ 
    \cline{2-6}
    & \textbf{Sem.} & \textbf{Syn.} & \textbf{All} & \textbf{Pr.} & \textbf{Sp.} \\ \hline \hline
    \verb|SISG(cj)|\textsuperscript{\dag} & \textit{.478} & \textit{.385} & \textit{.432} & \textit{-} & \textit{.677} \\
    \hline
    \verb|SG| & .423 & .495 & .459 & .608 & .627 \\
    \verb|SISG(c)| & .450 & .591 & .520 & .620 & .612 \\
    \verb|SISG(cj)| & .398 & .484 & .441 & \textbf{.665} & .671 \\
    \verb|SISG(cj)|\textsuperscript{\ddag} & .414 & .487 & .451 & .654 & \textbf{.674} \\
    \hline
    \verb|SISG(cjh3)| & \textbf{.340} & \textbf{.450} & \textbf{.395} & .634 & .633 \\
    \verb|SISG(cjh4)| & .349 & .456 & .402 & .624 & .617 \\
    \verb|SISG(cjhr)| & .355 & .462 & .409 & .650 & .647 \\
    \hline
    \end{tabular}
    \begin{tablenotes}
		\item[\dag]{\footnotesize reported by \citet{park2018subword}}
		\newline
		\item[\ddag]{\footnotesize pre-trained embeddings provided by the authors of \citet{park2018subword} run with our evaluation script}
	\end{tablenotes}
\end{threeparttable}
\end{center}
\caption{\label{tab:analogyandsim} Word analogy and similarity evaluation results. For word analogy test, the table shows cosine distances averaged over semantic and syntactic word pairs (lower is better). For word similarity test, we report on Pearson and Spearman correlations between predictions and ground truth scores. We observe that our approach (SISG(cjh)) shows significant improvements in the analogy test but some deterioration in the similarity test. Note that the evaluation results reported by the original authors (\dag) are largely different from our results. This might be due to differences in implementation details, hence we report and compare with the results of the authors' embeddings run on our test script ($\ddag$).}
\end{table}

\subsection{Intrinsic Evaluation}

\subsubsection{Word Analogy Test} Following the experimental protocol of \citet{park2018subword}, given an analogy pair $a:b \leftrightarrow c:d$, we compute the cosine distance between the word vectors $\mathbf{v}_a + \mathbf{v}_b - \mathbf{v}_c$ and $\mathbf{v}_d$ using the following equation: $1 - \cos(x, y)$, where $\cos$ is the cosine similarity. The Korean word analogy dataset \cite{park2018subword} consists of 10,000 quadruples, designed with a set of semantic and syntactic features specific to the Korean. We report our findings in Table \ref{tab:analogyandsim}. We observe that our approach improves significantly compared to the previous state-of-the-art models in both semantic and syntactic word analogy detection.

\subsubsection{Word Similarity Test}
The word similarity test aims to evaluate the correlation between word vector distances and human-annotated scores. 
We use the Korean WS353 dataset provided by \citet{park2018subword}.
Our results are shown in Table \ref{tab:analogyandsim}. Although our approaches do not outperform the jamo-level baseline (\verb|SISG(cj)|), we observe that Hanja-level information provides some advantage compared to just using pure character-level n-grams (\verb|SISG(c)|). 
Also note that WS353 is a relatively small dataset, which could be prone to statistical bias.

\subsubsection{Analysis} 
How much of the improvement can be explained by transfer learning from Chinese? Word analogy test results show that word representations trained with pre-trained Chinese n-grams perform better than those trained without (\verb|SISG(cjhr)|), supporting our claim that our approach is able to transfer relevant knowledge from the Chinese language for detecting analogical relationships. However, for the word similarity test, word vectors trained without Chinese embeddings perform better, suggesting that there are some trade-offs.

\subsection{Downstream Tasks}
In this section, we try to demonstrate the effectiveness of our approach to downstream tasks with two specific cases.

\subsubsection{Korean News Headline Generation}
To show that our approach helps supervised models in gaining deeper understanding of Korean texts, we devise a new task using news articles, which requires an understanding of the semantics of Korean texts. 
We collected Korean news articles published from 2017 January to February. 
The dataset covers balanced categories (e.g. politics, sports, world, etc.) and contains the headline title as the annotation, akin to the CNN/Daily Mail Dataset \cite{hermann2015teaching}. 
The dataset is relatively large - containing 840,205 news article and title pairs.

We use the encoder-decoder architecture supported by OpenNMT \cite{opennmt} framework for the task, where the encoder is a bidirectional LSTM cell and the decoder is an LSTM cell (the hidden size is 512 for both) with a bridging feed-forward layer between the two RNNs. 
We employ soft attention \cite{bahdanau2014neural} to generate news headlines given the first three sentences of an article body. We tokenize each headline using a Korean morpheme tokenizer\footnote{\href{https://github.com/shin285/KOMORAN}{https://github.com/shin285/KOMORAN}}
such that decoder tokens are relatively dense. 
The copy mechanism \cite{gu2016incorporating}, a popular technique of sequence-to-sequence models for translation and summarization, cannot be applied directly to our model, as the encoder tokens (space tokenized) and the decoder tokens (morpheme tokenized) do not share the same token space.
Word vectors obtained from both baselines and our approach are used to initialize the encoder embeddings before the training begins. The dataset is split into training, validation, and test sets (8:1:1). We report performances on the test set after validating our training models on the validation set. We evaluate the titles generated by our models with the BLEU \cite{papineni2002bleu}\footnote{Although ROUGE is a more widely adopted measure in English literature, no equivalent exists for Korean.}. 
We also report per-word perplexity obtained from the models.

As shown in Table \ref{tab:headline}, models trained with our embeddings perform better and have better language modeling capability than those trained with the previous state-of-the-art embeddings. 
This improvement can be partially explained by the fact that formal Korean texts, such as news articles, are more likely to contain Sino-Korean words, allowing models with awareness of Chinese to gain an edge in Korean understanding.

\begin{table}[t!]
\small
\begin{center}
\begin{threeparttable}
    \begin{tabular}{|l|cccc|c|}
    \hline \multirow{2}{*}{\textbf{Embeddings}} & \multicolumn{4}{c|}{\textbf{BLEU}} & \multirow{2}{*}{\textbf{PPL}}\\
    \cline{2-5}
    & \textbf{1} & \textbf{2} & \textbf{3} & \textbf{4} & \\ 
    \hline \hline
    None & 26.02 & 7.76 & 3.08 & 1.38 & 5.335\\
    \hline
    \verb|SG| & 30.33 & 10.20 & 4.29 & 1.98 & 4.122 \\
    \verb|SISG(c)| & 31.34 & 10.96 & 4.69 & 2.19 & 3.942 \\
    \verb|SISG(cj)| & 31.78 & 11.17 & 4.80 & 2.25 & 3.938 \\
    \verb|SISG(cj)|\textsuperscript{\dag} & 31.77 & 11.16 & 4.81 & 2.27 & 3.940 \\
    \hline
    \verb|SISG(cjh3)| & \textbf{32.03} & 11.25 & 4.83 & 2.27 & 3.941 \\
    \verb|SISG(cjh4)| & 32.02 & \textbf{11.34} & \textbf{4.92} & \textbf{2.30} & \textbf{3.909} \\
    \hline
    \end{tabular}
    \begin{tablenotes}
        \item[\dag]{\footnotesize pre-trained embeddings by \citet{park2018subword}}
    \end{tablenotes}
\end{threeparttable}
\end{center}
\caption{\label{tab:headline} Korean News Headline Generation results. }
\end{table}

\subsubsection{Sentiment Analysis}
To estimate the performance of our approach on a different task then the previous experiment, we conduct an experiment on Naver Sentiment Movie Corpus (NSMC)\footnote{\href{https://github.com/e9t/nsmc}{https://github.com/e9t/nsmc}}. 
This dataset consists of 200K movie reviews, each of which is labeled with its sentiment, i.e. positive or negative.
We split the corpus into training (100K), validation (50K), and test sets (50K). 
It is worth noting that this case study is designed to figure out to what extent our embeddings are generally applicable for even cases, where most of the sentences in the dataset consist of spoken words rather than written words and often do not contain Sino-Korean words.

We utilize a basic LSTM \cite{hochreiter1997long} module as a sentence encoder to exclude possible exceptional gains from sophisticated encoders, concentrating on the usefulness of input word embeddings. Likewise the previous experiment, word vectors obtained from both baselines and our approach are used as input for the encoder.
We regard the last hidden state of the LSTM as the sentence representation, which is consumed by a feed-forward network followed by a softmax classifier. The dimension of the LSTM cells is fixed as 300.
Each result from the baselines and our models is the average of 3 independent runs initialized with different random seeds, and the outcome of each run is chosen by the performance (F1-score) on the validation set.

From Table \ref{tab:NSMC}, we confirm our approach is comparable to the best previous one and general enough to be employed in most of the downstream tasks, even though the performance of our approach is somewhat unsatisfactory.
While not strictly proven, we conjecture one possible reason for the unsatisfying performance is the low reliability of the Hanja tagger we leveraged, which is not guaranteed to work well with spoken language. Specifically, we have manually inspected the tagging results and observed that there are some errors which can lead to performance degradation.
This observation points out a limitation of our current approach, that is, the dependence on external taggers, encouraging us to develop an integrated approach, leaving as future work not resorting to the taggers.

\begin{table}[t!]
\small
\begin{center}
\begin{threeparttable}
    \begin{tabular}{|l|cccc|}
    \hline \multirow{2}{*}{\textbf{Embeddings}} & \multicolumn{4}{|c|}{\textbf{Model: LSTM}} \\
    \cline{2-5}
    & \textbf{Acc.} & \textbf{P} & \textbf{R} & \textbf{F1} \\
    \hline \hline
    \verb|SISG(c)| & 77.43 & 75.89 & 80.41 & 78.08 \\
    \verb|SISG(cj)| & \textbf{83.16} & 82.36 & \textbf{84.66} & \textbf{83.50} \\
    \hline
    \verb|SISG(cjh3)| & 81.61 & 81.23 & 82.28 & 81.75 \\
    \verb|SISG(cjh4)| & 82.25 & \textbf{82.57} & 81.77 & 82.17 \\
    \hline
    \end{tabular}
        \begin{tablenotes}
    \end{tablenotes}
\end{threeparttable}
\end{center}
\caption{\label{tab:NSMC} Naver Sentiment Movie Corpus results. We report accuracy (Acc.), precision (P), recall (R), and F1-score (F1) following \citet{park2018subword}. Each reported result is the average of several runs initialized with different random seeds.}
\end{table}

\section{Related Work}
Although the word is usually regarded as the smallest and basic unit for most NLP pipelines, there is a recent trend of utilizing subword information for enriching word representations (\citealt{sennrich2016neural}; \citealt{ bojanowski2017enriching}, to name a few), or considering subwords themselves as input directly for NLP models \cite{zhang2015character, kim2016character, peters2018deep, devlin2018bert}.

As Korean is agglutinative \cite{song2006korean}, the current literature in Korean word representations mainly focus on subword structures such as morphemes \cite{, edmiston2018compositional}, syllables \cite{choi2017syllable} and Jamo \cite{choi2016grapheme, stratos2017sub, park2018subword}.
We here move forward one step further by incorporating Hanja information explicitly together with the aforementioned  subword information.

On the other hand, cross-lingual representations is a trending topic in literature (\citealt{lample2018word}; \citealt{conneau2018xnli}; \citealt{lample2019cross}, to name a few).
Nevertheless, to the best of our knowledge, our work is the first to introduce character-level cross-lingual transfer learning based on etymological grounds. Furthermore, the novelty of our work lies of the fact that we use separated vocabulary instead of shared vocabulary such as Byte Pair Encoding (BPE) \cite{sennrich2016neural}.

\section{Conclusions}

We have presented a method of training Korean word representations with Hanja. 
In our extensive experiments, we have demonstrated that our approach is effective in infusing more semantics into Korean word embeddings.
One potential issue with our method, as already mentioned, is that it relies on external Hanja annotation, even though this can be mitigated to some extent by off-the-shelf taggers, as in this work. 
Thus, it would be an attractive direction to take as future work developing an end-to-end system.

\section*{Acknowledgments}
We thank Daniel Edmiston and anonymous reviewers for their helpful feedback. This work was supported by BK21 Plus for Pioneers in Innovative Computing (Dept. of Computer Science and Engineering, SNU) funded by the National Research Foundation of Korea (NRF) (21A20151113068).

\bibliography{2019-yookim-arxiv}
\bibliographystyle{acl_natbib}
\appendix 

\section{Supplemental Material}
\label{sec:supplemental}

\subsection{Implementation Details for SISG Model}

We trained our models over 5 epochs with the following parameters: 300 word vector dimensions, 20m bucket size to cover all n-grams including Hanja, 0.0001 sampling threshold, 0.05 learning rate, 5 negative sampling size, 3-5 jamo n-gram sizes, and 5 context window size. More details can be found at \url{https://github.com/kaniblu/hanja-sisg}. 

\subsection{Ablation Results}

In this supplementary material, we include results for different experimental settings such as the type of training corpus and the character n-gram sizes (Table \ref{tab:supp-ablation}).

\begin{table*}[t!]
\begin{center}
\begin{threeparttable}
    \begin{tabular}{|lcc|ccc|cc|}
    \hline \multirow{2}{*}{\textbf{Method}} & \multirow{2}{*}{\textbf{Dataset}} & \multirow{2}{*}{\textbf{Char. N-Grams}} & \multicolumn{3}{c|}{\textbf{Analogy}} & \multicolumn{2}{c|}{\textbf{Similarity}}\\ 
    \cline{4-8}
    &&& \textbf{Sem.} & \textbf{Syn.} & \textbf{All} & \textbf{Pearson} & \textbf{Spearman} \\ \hline \hline
    \verb|SISG(cj)|\textsuperscript{\dag} & Original & 1-6 & \textit{0.478} & \textit{0.385} & \textit{0.432} & \textit{-} & \textit{0.677} \\
    \hline
    \verb|SG| & Improved & - & 0.423 & 0.495 & 0.459 & 0.608 & 0.627 \\
    \verb|SISG(c)| & Improved & 1-6 & 0.450 & 0.591 & 0.520 & 0.620 & 0.612 \\
    \verb|SISG(cj)| & Improved & 1-6 & 0.398 & 0.484 & 0.441 & \textbf{0.665} & 0.671 \\
    \verb|SISG(cj)| & Improved & 1-4 & 0.400 & 0.485 & 0.442 & 0.640 & 0.637 \\
    \verb|SISG(cj)| & Original & 1-6 & 0.399 & 0.488 & 0.444 & 0.659 & 0.661 \\
    \verb|SISG(cj)| & Original & 1-4 & 0.398 & 0.483 & 0.441 & 0.652 & 0.653 \\
    \verb|SISG(cj)|\textsuperscript{\ddag} & Original & 1-6 & 0.414 & 0.487 & 0.451 & 0.654 & \textbf{0.674} \\
    \hline
    \verb|SISG(cjh3)| & Improved & 1-6 & \underline{0.340} & \textbf{0.450} & \textbf{0.395} & 0.634 & 0.633 \\
    \verb|SISG(cjh3)| & Improved & 1-4 & \textbf{0.339} & 0.453 & 0.396 & 0.612 & 0.605 \\
    \verb|SISG(cjh4)| & Improved & 1-6 & 0.349 & 0.456 & 0.402 & 0.624 & 0.617 \\
    \verb|SISG(cjh4)| & Improved & 1-4 & 0.349 & 0.456 & 0.402 & 0.640 & 0.638 \\
    \verb|SISG(cjhr)| & Improved & 1-6 & 0.355 & 0.462 & 0.409 & 0.650 & 0.647 \\
    \hline
    \end{tabular}
    \begin{tablenotes}
		\item[\dag]{\footnotesize reported by \citet{park2018subword}.}
		\newline
		\item[\ddag]{\footnotesize pre-trained embeddings provided by the authors of \citet{park2018subword} run with our evaluation script.}
	\end{tablenotes}
\end{threeparttable}
\end{center}
\cprotect\caption{\label{tab:supp-ablation} Full results of our ablation studies. By conducting experiments on varying character n-gram lengths, we determine that character n-grams ranging from 1-6 yield better results for our model. The dataset columns shows two different types of datasets: \verb|original| and \verb|improved|. The \verb|original| dataset is the corpus originally provided by the authors of \cite{park2018subword}, and the \verb|improved| dataset is the one that has been further data-cleansed from the original corpus. The results show that word vectors trained on the improved corpus achieve marginal but still meaningful improvement in quality.}
\end{table*}

\subsection{Full Results for Word Analogy}

Here we present the full results for the word analogy test (Table \ref{tab:supp-analogy}), including all categories of test analogy. 

\begin{table*}[t!]
\small
\begin{center}
\begin{threeparttable}
    \begin{tabular}{|l|ccccc|ccccc|c|}
    \hline \multirow{2}{*}{\textbf{Method}} & \multicolumn{5}{c|}{\textbf{Semantic}} & \multicolumn{5}{c|}{\textbf{Syntactic}} & \multirow{2}{*}{\textbf{All}}\\ 
    \cline{2-11}
    & \textbf{City} & \textbf{Sex} & \textbf{Name} & \textbf{Lang} & \textbf{Misc} & \textbf{Case} & \textbf{Tense} & \textbf{Voice} & \textbf{Form} & \textbf{Honor} & \\ \hline \hline
    \verb|SISG(cj)|\textsuperscript{\dag} & \textit{0.425} & \textit{0.498} & \textit{0.561} & \textit{0.354} & \textit{0.554} & \textit{0.210} & \textit{0.414} & \textit{0.426} & \textit{0.507} & \textit{0.367} & \textit{0.432} \\
    \hline
    \verb|SG| & 0.471 & 0.478 & 0.413 & 0.338 & 0.419 & 0.540 & 0.482 & 0.517 & 0.486 & 0.449 & 0.459 \\
    \verb|SISG(c)| & 0.492 & 0.512 & 0.436 & 0.401 & 0.408 & 0.645 & 0.573 & 0.597 & 0.554 & 0.584 & 0.520 \\
    \verb|SISG(cj)| & 0.430 & 0.466 & 0.384 & 0.331 & 0.377 & 0.591 & 0.473 & 0.485 & 0.489 & 0.384 & 0.441 \\
    \verb|SISG(cj)|\textsuperscript{\ddag} & 0.449 & 0.468 & 0.400 & 0.341 & 0.412 & 0.576 & 0.479 & 0.485 & 0.484 & 0.413 & 0.451 \\
    \hline
    \verb|SISG(cjh3)| & \textbf{0.363} & \textbf{0.424} & \textbf{0.326} & \textbf{0.258} & \textbf{0.328} & \textbf{0.558} & \textbf{0.439} & \textbf{0.461} & \textbf{0.444} & \textbf{0.348} & \textbf{0.395} \\
    \verb|SISG(cjh4)| & 0.377 & 0.423 & 0.333 & 0.270 & 0.340 & 0.563 & 0.448 & 0.463 & 0.457 & 0.351 & 0.402 \\
    \verb|SISG(cjhr)| & 0.389 & 0.432 & 0.343 & 0.274 & 0.338 & 0.569 & 0.449 & 0.468 & 0.466 & 0.355 & 0.409 \\
    \hline
    \end{tabular}
    \begin{tablenotes}
		\item[\dag]{\footnotesize reported by \citet{park2018subword}}
		\newline
		\item[\ddag]{\footnotesize pre-trained embeddings provided by the authors of \citet{park2018subword} run with our evaluation script}
	\end{tablenotes}
\end{threeparttable}
\end{center}
\caption{\label{tab:supp-analogy} Full results on the word analogy test. Note that the evaluation results reported by the original authors (\dag) is largely different from our results. This might be due to differences in implementation details, hence we report and compare only with the results of the authors' embeddings run on our test script ($\ddag$).}
\end{table*}

\end{document}